\documentclass[11pt,a4paper]{article}
\usepackage{authblk}
\usepackage[hang,flushmargin]{footmisc}

\usepackage[hyperref]{naaclhlt2018}
\usepackage{times}
\usepackage{latexsym}
\usepackage[margin=1in]{geometry} 
\usepackage{url}
\usepackage{amsmath,amsthm,amsfonts}
\usepackage{booktabs}
\usepackage{graphicx}
\usepackage{array,colortbl,multirow,multicol,ctable}
\usepackage{xcolor}
\usepackage{amssymb}

\usepackage{caption}

\title{Dense Information Flow for Neural Machine Translation}

\author[1]{Yanyao Shen}
\author[2]{Xu Tan}
\author[3]{Di He}
\author[2]{Tao Qin}
\author[2]{Tie-Yan Liu}
\affil[1]{University of Texas at Austin}
\affil[2]{Microsoft Research, Asia}
\affil[3]{Key Laboratory of Machine Perception, MOE, School of EECS, Peking University}
\affil[ ]{shenyanyao@utexas.edu, \{xuta,taoqin,tie-yan.liu\}@microsoft.com,di\_he@pku.edu.cn}

\aclfinalcopy 
\def\aclpaperid{209} 

\begin{document}

\maketitle

\begin{abstract}
Recently, neural machine translation has achieved remarkable progress by introducing well-designed deep neural networks into its encoder-decoder framework.
From the optimization perspective, 
residual connections are adopted to improve learning performance for both encoder and decoder in most of these deep architectures, and advanced attention connections are applied as well. 
Inspired by the success of the DenseNet model in computer vision problems, in this paper, we propose a densely connected NMT architecture (DenseNMT) that is able to train more efficiently for NMT. 
The proposed DenseNMT not only allows dense connection in creating new features for both encoder and decoder, but also uses the dense attention structure to improve attention quality. 
Our experiments on multiple datasets show that  DenseNMT structure is more competitive and efficient~\footnote{Code available at:  https://github.com/yanyao-shen/fairseq}. 
\end{abstract}

\section{Introduction}
Neural machine translation (NMT) is a challenging task that attracts lots of attention in recent years. 
Starting from the encoder-decoder framework~\cite{cho2014learning}, NMT starts to show promising results in many language pairs.
The evolving structures of NMT models in recent years have made them achieve higher scores and become more favorable. 
The attention mechanism~\cite{bahdanau2014neural} added on top of encoder-decoder framework is shown to be very useful to automatically find alignment structure, and single-layer RNN-based structure has evolved into deeper models with more efficient transformation functions~\cite{convs2s,slicenet,transformer}.

One major challenge of NMT is that its models are hard to train in general due to the complexity of both the deep models and languages. 
From the optimization perspective, deeper models are hard to efficiently back-propagate the gradients, and this phenomenon as well as its solution is better explored in the computer vision society. 
Residual networks (ResNet)~\cite{resnet} achieve great performance in a wide range of tasks, including image classification and image segmentation. 
Residual connections allow features from previous layers to be accumulated to the next layer easily, and make the optimization of the model efficiently focus on refining upper layer features. 

NMT is considered as a challenging problem due to its sequence-to-sequence generation framework, and the goal of comprehension and reorganizing from one language to the other. 
Apart from the encoder block that works as a feature generator, the decoder network combining with the attention mechanism bring new challenges to the optimization of the models. 
While nowadays best-performing NMT systems use residual connections, we question whether this is the most efficient way to propagate information through deep models. 
In this paper, inspired by the idea of using dense connections for training computer vision tasks~\cite{densenet}, we propose a densely connected NMT framework (DenseNMT) that efficiently propagates information from the encoder to the decoder through the attention component. 
Taking the CNN-based deep architecture as an example, we verify the efficiency of DenseNMT. Our contributions in this work include:
(i) by comparing the loss curve, we show that DenseNMT allows the model to pass information more efficiently, and speeds up training; 
(ii) we show through ablation study that dense connections in all three blocks altogether help improve the performance, while not increasing the number of parameters; 
(iii) DenseNMT allows the models to achieve similar performance with much smaller embedding size; 
(iv) DenseNMT on IWSLT14 German-English and Turkish-English translation tasks achieves new benchmark BLEU scores, and the result on WMT14 English-German task is more competitive than the residual connections based baseline model.

\section{Related Work}
\paragraph{ResNet and DenseNet. }
ResNet~\cite{resnet} proposes residual connections, which directly add representation from the previous layer to the next layer. Originally proposed for image classification tasks, the residual structure have proved its efficiency in model training across a wide range of tasks, and are widely adopted in recent advanced NMT models~\cite{gnmt,transformer,convs2s}. 
Following the idea of ResNet, DenseNet~\cite{densenet} further improves the structure and achieves state-of-the-art results. It allows the transformations (e.g., CNN) to be directly calculated over all previous layers. The benefit of DenseNet is to encourage upper layers to create new representations instead of refining the previous ones. 
On other tasks such as segmentation, dense connections also achieve high performance~\cite{jegou2017one}. 
Very recently, \cite{godin2017improving} shows that dense connections help improve language modeling as well. 
Our work is the first to explore dense connections for NMT tasks.

\paragraph{Attention mechanisms in NMT.}
The attention block is proven to help improve inference quality due to existence of alignment information~\cite{bahdanau2014neural}. 
Traditional sequence-to-sequence architectures~\cite{kalchbrenner2013recurrent,cho2014learning} pass the last hidden state from the encoder to the decoder; hence source sentences of different length are encoded into a fixed-size vector (i.e., the last hidden state), and the decoder should catch all the information from the vector. 
Later, early attention-based NMT architectures, including ~\cite{bahdanau2014neural}, pass all the hidden states (instead of the last state) of the last encoder layer to the decoder. The decoder then uses an attention mechanism to selectively focus on those hidden states while generating each word in the target sentence.
Latest architecture~\cite{convs2s} uses multi-step attention, which allows each decoder layer to acquire separate attention representations, in order to maintain different levels of semantic meaning. %
They also enhance the performance by using embeddings of input sentences. 
In this work, we further allow every encoder layer to directly pass the information to the decoder side. 

\paragraph{Encoder/decoder networks.}
RNNs such as long short term memory (LSTM) are widely used in NMT due to their ability of modeling long-term dependencies. 
Recently, other more efficient structures have been proposed in substitution for RNN-based structures, which includes convolution~\cite{convs2s,slicenet} and self-attention~\cite{transformer}. 
More specifically, ConvS2S~\cite{convs2s} uses convolution filter with a gated linear unit, Transformer~\cite{transformer} uses self-attention function before a two-layer position-wise feed-forward networks, and SliceNet~\cite{slicenet} uses a combination of ReLU, depthwise separable convolution, and layer normalization.  
The advantage of these non-sequential transformations is the significant parallel speedup as well as more advanced performances, which is the reason we select CNN-based models for our experiments.

\section{DenseNMT}

\begin{figure}[t]
\centering 
\small 
\captionsetup{font=small}
\includegraphics[width=0.32\textwidth]{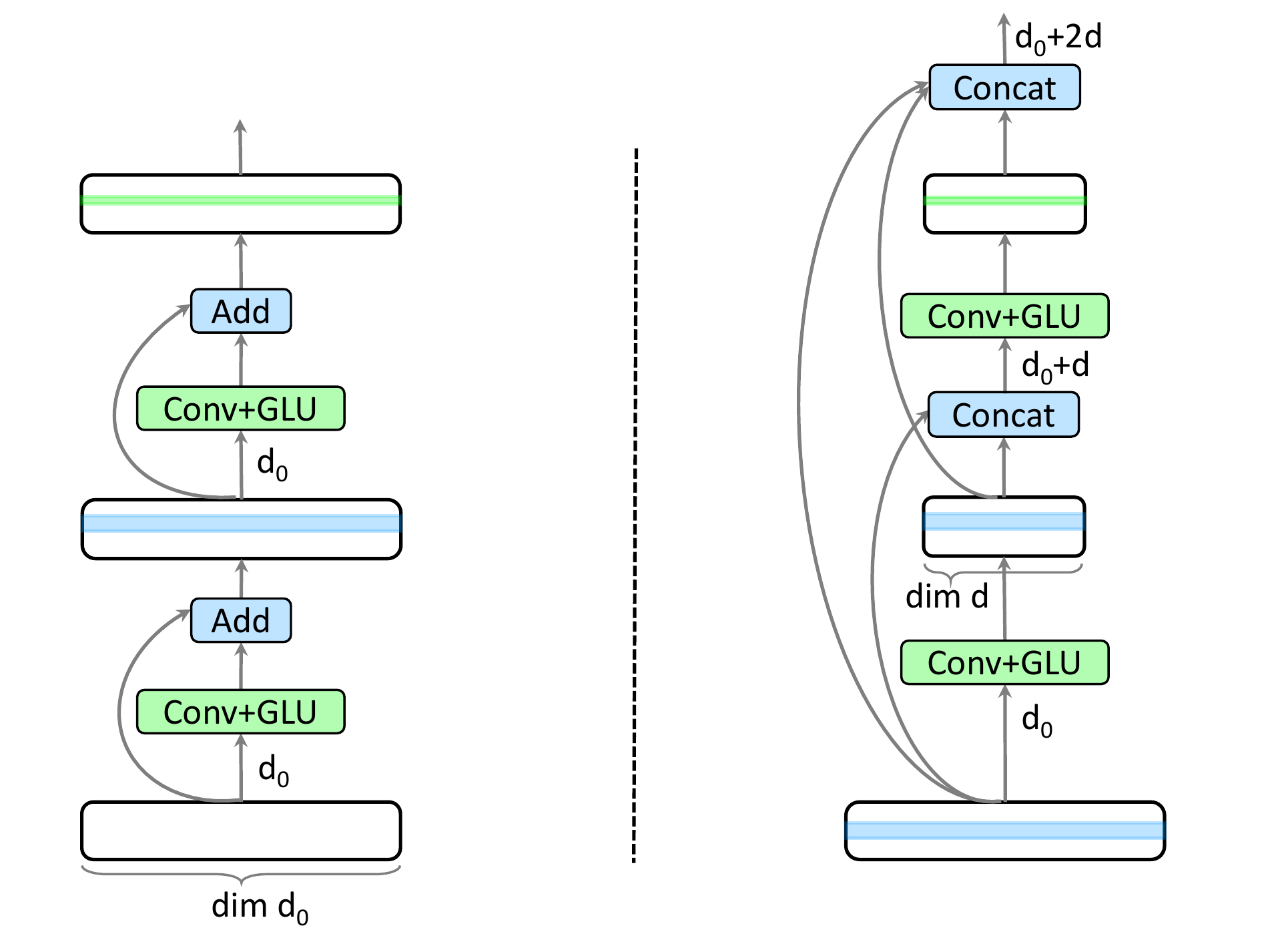}
\caption{Comparison of dense-connected encoder and residual-connected encoder. Left: regular residual-connected encoder. Right: dense-connected encoder. Information is directly passed from blue blocks to the green block.}
\label{fig:enc}
\end{figure}

\begin{figure*}[t]
\centering 
\captionsetup{font=small}
\includegraphics[width=1\textwidth]{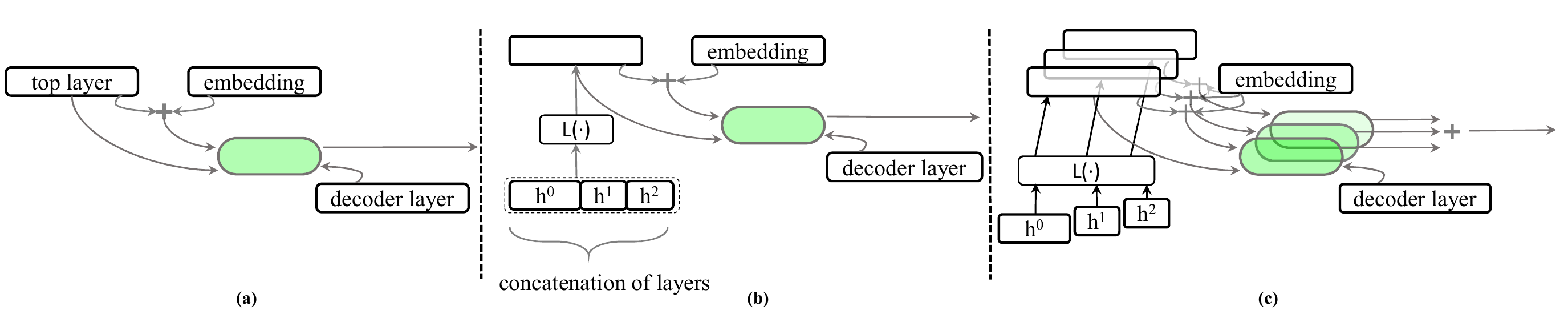}
\caption{Illustration of DenseAtt mechanisms. For clarity, We only plot the attention block for a single decoder layer. (a): multi-step attention~\cite{convs2s}, (b): DenseAtt-1, (c): DenseAtt-2. $\mathcal{L}(\cdot)$ is the linear projection function. The ellipsoid stands for the core attention operation as shown in Eq. (\ref{eqt:att}).}
\label{fig:att}
\end{figure*}

In this section, we introduce our DenseNMT architecture. 
In general, compared with residual connected NMT models, DenseNMT allows each layer to provide its information to all subsequent layers directly. 
Figure \ref{fig:enc}-\ref{fig:att} show the design of our model structure by parts. 

We start with the formulation of a regular NMT model. 
Given a set of sentence pairs $S\!=\!\{({x^i},{y^i})|i\!=\!1,\!\cdots,\! N\}$, an NMT model learns parameter $\theta$ by maximizing the log-likelihood function:
\begin{equation}
\sum_{i=1}^{N} \log \mathcal{P}({y^i}|{x^i};\theta). 
\end{equation}
For every sentence pair $(x,y)\in S$,  $\mathcal{P}({y}|{x};\theta)$ is calculated based on the decomposition:
\begin{equation}
\mathcal{P}({y}|{x};\theta) = \prod_{j=1}^{m}\mathcal{P}(y_j|{y_{<j}},{x};\theta),
\end{equation}
where $m$ is the length of sentence $y$. Typically, NMT models use the encoder-attention-decoder framework~\cite{bahdanau2014neural}, and potentially use multi-layer structure for both encoder and decoder. Given a source sentence $x$ with length $n$, the encoder calculates hidden representations by layer.
We denote the representation in the $l$-th layer as $h^l$, with dimension $n\times d^l$, where $d^l$ is the dimension of features in layer $l$. 
The hidden representation at each position $h_j^l$ is either calculated by:
\begin{equation}
h_j^l = \mathcal{H}^{\mathtt{rec}}(h_j^{l-1}, h_{j-1}^l)
\end{equation}
for recurrent transformation $\mathcal{H}^{\mathtt{rec}}(\cdot)$ such as LSTM and GRU, or by:
\begin{equation}
h_j^l = \mathcal{H}^{\mathtt{par}}(h^{l-1})
\end{equation}
for parallel transformation $\mathcal{H}^{\mathtt{par}}(\cdot)$. 
On the other hand, the decoder layers $\{z^l\}$ follow similar structure, while getting extra representations from the encoder side. 
These extra representations are also called \textit{attention}, and are especially useful for capturing alignment information.

In our experiments, we use convolution based transformation for $\mathcal{H}^{\mathtt{par}}(\cdot)$ due to both its efficiency and high performance, more formally,
\begin{equation}\label{eqt:hfunc}
h_j^l = \mathtt{GLU}([h_{j-r}^{l-1}, \cdots, h_{j+r}^{l-1}] W^{l} + b^l)\triangleq \mathcal{H}(h^{l-1}).
\end{equation}
$\mathtt{GLU}$ is the gated linear unit proposed in~\cite{dauphin2016language} and the kernel size is $2r+1$. 
DenseNMT is agnostic to the transformation function, and we expect it to also work well combining with other transformations, such as LSTM, self-attention and depthwise separable convolution. 

\subsection{Dense encoder and decoder}

Different from residual connections, later layers in the dense encoder are able to use features from all previous layers by concatenating them: 
\begin{equation}\label{eqt:dense}
h^{l+1} = \mathcal{H}([h^{l}, h^{l-1},\cdots, h^0]).
\end{equation}
Here, $\mathcal{H}(\cdot)$ is defined in Eq. (\ref{eqt:hfunc}), $[\cdot]$ represents concatenation operation. 
Although this brings extra connections to the network, with smaller number of features per layer, the architecture encourages feature reuse, and can be more compact and expressive. 
As shown in Figure \ref{fig:enc}, when designing the model, the hidden size in each layer is much smaller than the hidden size of the corresponding layer in the residual-connected model. 

\begin{figure}[!t]
\centering 
\small
\captionsetup{font=small}
\includegraphics[width=0.4\textwidth]{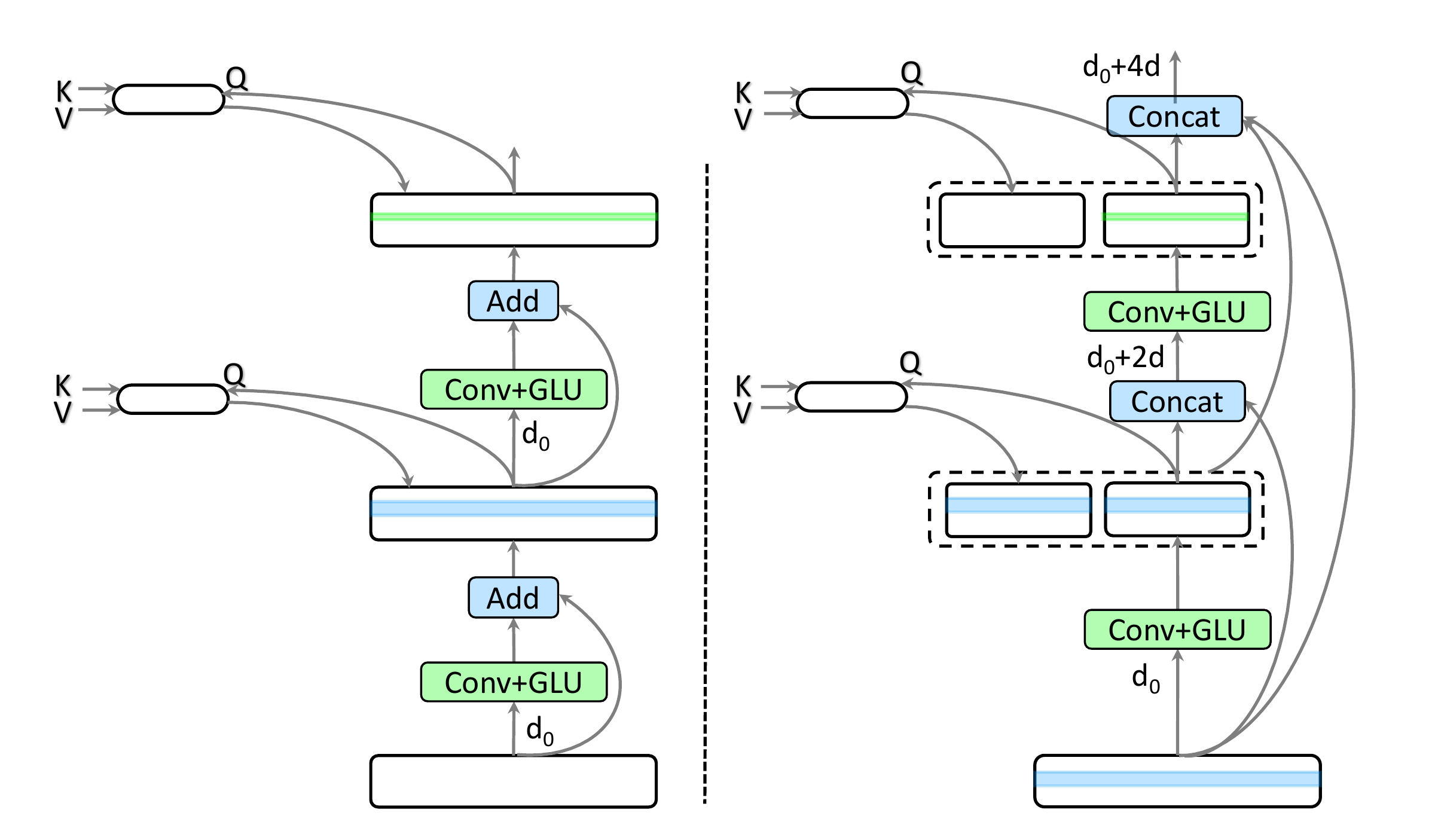}
\caption{Comparison of dense-connected decoder and residual-connected decoder. Left: regular residual-connected decoder. Right: dense-connected decoder. Ellipsoid stands for attention block. Information is directly passed from blue blocks to the green block. }
\label{fig:dec}
\end{figure} 

While each encoder layer perceives information from its previous layers, each decoder layer $z^{l+1}$ has two information sources: previous layers $z^{i}, i\le l$, and attention values $a^{i}, i\le l$. Therefore, in order to allow dense information flow, we redefine the generation of $(l+1)$-th layer as a nonlinear function over all its previous decoder layers and previous attentions. This can be written as: 
\begin{equation}
z^{l+1} = \mathcal{H}([z^{l}, a^{l}, z^{l-1}, a^{l-1},\cdots, z^1, a^1, z^0]),
\end{equation}
where $a^i$ is the attention value using $i$-th decoder layer and information from encoder side, which will be specified later. 
Figure \ref{fig:dec} shows the comparison of a dense decoder with a regular residual decoder. The dimensions of both attention values and hidden layers are chosen with smaller values, yet the perceived information for each layer consists of a higher dimension vector with more representation power.
The output of the decoder is a linear transformation of the concatenation of all layers by default. To compromise to the increment of dimensions, we use summary layers, which will be introduced in Section 3.3. With summary layers, the output of the decoder is only a linear transformation of the concatenation of the upper few  layers.

\subsection{Dense attention}

Prior works show a trend of designing more expressive attention mechanisms (as discussed in Section 2). However, most of them only use the last encoder layer.
In order to pass more abundant information from the encoder side to the decoder side, the attention block needs to be more expressive.
Following the recent development of designing attention architectures, we propose DenseAtt as the dense attention block, which serves for the dense connection between the encoder and the decoder side. 
More specifically, two options are proposed accordingly. 
For each decoding step in the corresponding decoder layer, the two options both calculate attention using multiple encoder layers. The first option is more compressed, while the second option is more expressive and flexible. We name them as DenseAtt-1 and DenseAtt-2 respectively. Figure \ref{fig:att}  shows the architecture of (a) multi-step attention~\cite{convs2s}, (b) DenseAtt-1, and (c) DenseAtt-2 in order. 
In general, a popular multiplicative attention module can be written as:
\begin{equation}\label{eqt:att}
\mathcal{F}\left(Q,K,V\right) = \mathtt{Softmax}\left(Q\times K\right)\times V,
\end{equation}
where $Q,K,V$ represent query, key, value respectively. We will use this function $\mathcal{F}$ in the following descriptions.

\paragraph{DenseAtt-1}
In the decoding phase, we use a layer-wise attention mechanism, such that each decoder layer absorbs different attention information to adjust its output. Instead of treating the last hidden layer as the encoder's output, we treat the concatenation of all hidden layers from encoder side as the output. The decoder layer multiplies with the encoder output to obtain the attention weights, which is then multiplied by a linear combination of the encoder output and the sentence embedding. The attention output of each layer $a^l$ can be formally written as:
\begin{equation} \label{eqt:msmla-1}
a^{l} = \mathcal{F}\left(\!\mathcal{L}(z^l), \!\mathcal{L}\left( [\{h^i\}] \right), \!\mathcal{L}\left([\{h^i\}] \right)\!+\!\mathcal{L}(h^0)\right),
\end{equation}
where $\mathcal{F}(\cdot,\cdot,\cdot)$ is the multiplicative attention function, $[\cdot]$ is a concatenation operation that combines all features, and $\mathcal{L}(\cdot)$ is a linear transformation function that maps each variable to a fixed dimension in order to calculate the attention value. Notice that we explicitly write the $\mathcal{L}(h^0)$ term in (\ref{eqt:msmla-1}) to keep consistent with the multi-step attention mechanism, as pictorially shown in Figure \ref{fig:att}(a).

\paragraph{DenseAtt-2}
Notice that the transformation $\mathcal{L}([\{h^i\}])$ in DenseAtt-1 forces the encoder layers to be mixed before doing attention. 
Since we use multiple hidden layers from the encoder side to get an attention value, we can alternatively calculate multiple attention values before concatenating them. 
In another word, the decoder layer can get different attention values from different encoder layers. 
This can be formally expressed as: 
\begin{equation} \label{eqt:msmla-2}
a^{l} = \sum_{i=1}^{L} \mathcal{F}\left(\mathcal{L}(z^l), \mathcal{L}(h^i), \mathcal{L}([h^i,h^0])\right),
\end{equation}
where the only difference from  Eq. (\ref{eqt:msmla-1}) is that the concatenation operation is substituted by a summation operation, and is put after the attention function $\mathcal{F}$. 
This method further increases the representation power in the attention block, while maintaining the same number of parameters in the model.

\subsection{Summary layers}
Since the number of features fed into nonlinear operation is accumulated along the path, the parameter size increases accordingly. 
For example, for the $L$-th encoder  layer, the input dimension of features is  $(L-1)d+d_0$ , where $d$ is the feature dimension in previous layers, $d_0$ is the embedding size. 
In order to avoid the calculation bottleneck for later layers due to large $L$, we introduce the \textit{summary layer} for deeper models.
It summarizes the features for all previous layers and projects back to the embedding size, so that later layers of both the encoder and the decoder side do not need to look back further.
The summary layers can be considered as contextualized word vectors in a given sentence~\cite{mccann2017learned}. 
We add one summary layer after every $(\mathtt{sumlen}-1)$ layers, where $\mathtt{sumlen}$ is the hyperparameter we introduce. 
Accordingly, the input dimension of features is at most $(\mathtt{sumlen}-1)\cdot d + d_0$  for the last layer of the encoder.
Moreover, combined with the summary layer setting, our DenseAtt mechanism allows each decoder layer to calculate the attention value focusing on the last few encoder layers, which consists of the last contextual embedding layer and several dense connected layers with low dimension. In practice, we set $\mathtt{sumlen}$ as $5$ or $6$.

\subsection{Analysis of information flow}
Figure \ref{fig:enc} and Figure \ref{fig:dec} show the difference of information flow compared with a residual-based encoder/decoder. For residual-based models, each layer can absorb a single high-dimensional vector from its previous layer as the only information, while for DenseNMT, each layer can utilize several low-dimensional vectors  from its previous layers and a high-dimensional vector from the first layer (embedding layer) as its information. 
In DenseNMT, each layer directly provides information to its later layers. 
Therefore, the structure allows feature reuse, and encourages upper layers to focus on creating new features. 
Furthermore, the attention block allows the embedding vectors (as well as other hidden layers) to guide the decoder's generation more directly; therefore, during back-propagation, the gradient information can be passed directly to all encoder layers simultaneously.

\section{Experimental Setup}

\subsection{Datasets}

We use three datasets for our experiments: IWSLT14 German-English, Turkish-English, and WMT14 English-German. 

We preprocess the IWSLT14 German-English dataset following byte-pair-encoding (BPE) method~\cite{subword}\footnote{https://github.com/rsennrich/subword-nmt}. We learn 25k BPE codes using the joint corpus of source and target languages. 
We randomly select 7k from IWSLT14 German-English  as the development set
, and the test set is a concatenation of dev2010, tst2010, tst2011 and tst2012, which is widely used in prior works~\cite{ranzato2015sequence,bahdanau2016actor,msrphrase}.

For the Turkish-English translation task, we use the data provided by IWSLT14~\cite{cettolo2014report} and the SETimes corpus~\cite{cettolo2014report} following~\cite{sennrich2015improving}. After removing sentence pairs with  length ratio over 9, we obtain 360k sentence pairs. Since there is little commonality between the two languages, we learn 30k size BPE codes separately for Turkish and English. 
In addition to this, we give another preprocessing for Turkish sentences and use word-level English corpus. 
For Turkish sentences, following~\cite{gulcehre2015using,sennrich2015improving}, we use the morphology tool Zemberek with disambiguation by the morphological analysis~\cite{sak2007morphological} and removal of non-surface tokens\footnote{github.com/orhanf/zemberekMorphTR}. 
Following~\cite{sennrich2015improving}, we concatenate tst2011, tst2012, tst2013, tst2014 as our test set. We concatenate dev2010 and tst2010 as the development set.

We preprocess the WMT14 English-German\footnote{https://nlp.stanford.edu/projects/nmt/} dataset using a BPE code size of 40k. 
We use the concatenation of newstest2013 and newstest2012 as the development set.

\subsection{Model and architect design}
As the baseline model (\emph{BASE-4L}) for IWSLT14 German-English and Turkish-English, we use a 4-layer encoder, 4-layer decoder, residual-connected model\footnote{https://github.com/facebookresearch/fairseq}, with embedding and hidden size set as $256$ by default. 
As a comparison, we design a densely connected model with same number of layers, but the hidden size is set as $128$ in order to keep the model size consistent. The models adopting DenseAtt-1, DenseAtt-2 are named as \emph{DenseNMT-4L-1} and \emph{DenseNMT-4L-2} respectively.
In order to check the effect of dense connections on deeper models, we also construct a series of 8-layer models. We set the hidden number to be $192$, such that both 4-layer models and 8-layer models have similar number of parameters. 
For dense structured models, we set the dimension of hidden states to be $96$. 

\begin{figure}[t]
\centering 
\captionsetup{font=small}
\includegraphics[width=0.4\textwidth]{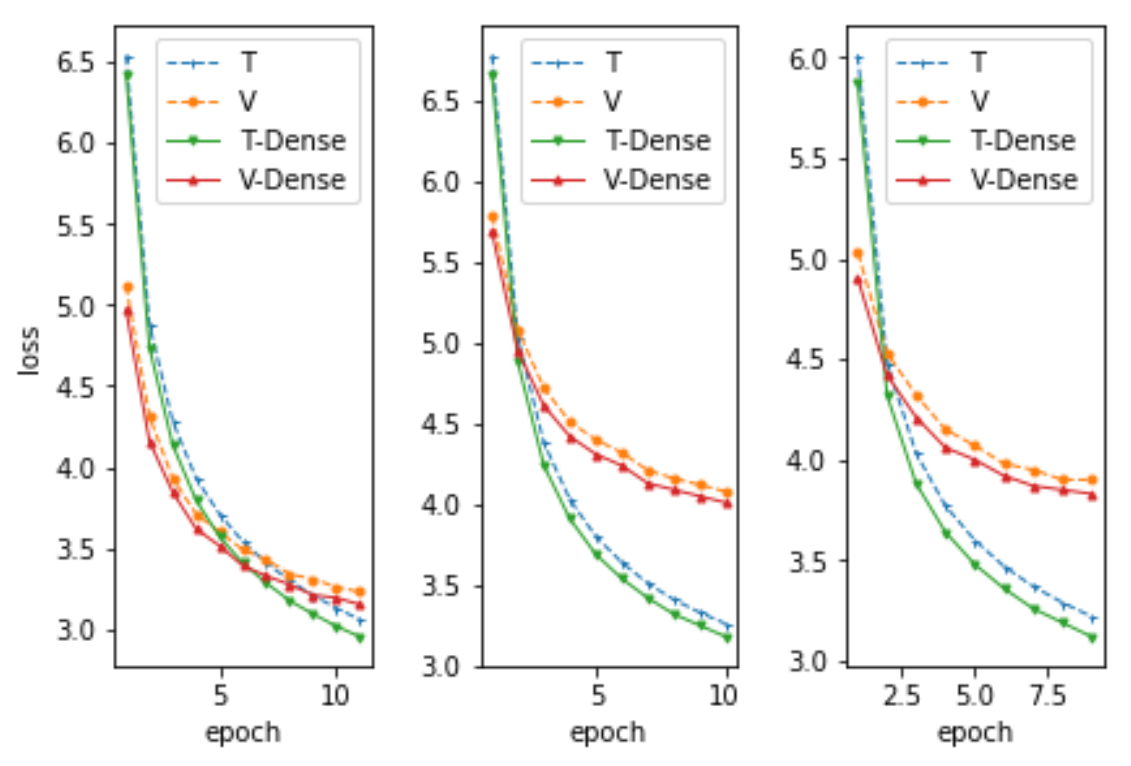}
\caption{Training curve (T) and validation curve (V) comparison. Left: IWSLT14 German-English (De-En). Middle: Turkish-English, BPE encoding (Tr-En). Right: Turkish-English, morphology encoding (Tr-En-morph).}
\label{fig:curve-appendix}
\end{figure}

Since NMT model usually allocates a large proportion of its parameters to the  source/target sentence embedding and softmax matrix, we explore in our experiments to what extent decreasing the dimensions of the three parts would harm the BLEU score. 
We change the dimensions of the source embedding, the target embedding as well as the softmax matrix simultaneously to smaller values, and then project each word back to the original embedding dimension through a linear transformation. 
This significantly reduces the number of total parameters, while not influencing the upper layer structure of the model. 

We also introduce three additional models we use for ablation study, all using 4-layer structure. 
Based on the residual connected \emph{BASE-4L} model, (1) \emph{DenseENC-4L} only makes encoder side dense, (2) \emph{DenseDEC-4L} only makes decoder side dense, and (3) \emph{DenseAtt-4L} only makes the attention dense using DenseAtt-2. There is no summary layer in the models, and both \emph{DenseENC-4L} and \emph{DenseDEC-4L} use hidden size $128$. Again, by reducing the hidden size, we ensure that different $4$-layer models have similar model sizes.

Our design for the WMT14 English-German model follows the best performance model provided in~\cite{convs2s}. 
The construction of our model is straightforward: our 15-layer model \emph{DenseNMT-En-De-15} uses dense connection with DenseAtt-2, $\texttt{sumlen}=6$. The hidden number in each layer is $1/4$ that of the original model, while the kernel size maintains the same. 

\subsection{Training setting}
We use Nesterov Accelerated Gradient (NAG)~\cite{nesterov1983method} as our optimizer, and the initial learning rate is set to $0.25$. For German-English and Turkish-English experiments, the learning rate will shrink by $10$ every time the validation loss increases. For the English-German dataset, in consistent with~\cite{convs2s}, the learning rate will shrink by $10$ every epoch since the first increment of validation loss. The system stops training until the learning rate is less than $10^{-4}$. 
All models are trained end-to-end without any warmstart techniques. 
We set our batch size for the WMT14 English-German dataset to be $48$, and additionally tune the length penalty parameter, in consistent with~\cite{convs2s}. For other datasets, we set batch size to be $32$.
During inference, we use a beam size of 5.

\section{Results}

\begin{table*}[t]
\centering
\small
\captionsetup{font=small}
\begin{tabular}{llccccccccccc}
\toprule
& & \multicolumn{3}{c}{\textbf{De-En}}  &  & \multicolumn{3}{c}{\textbf{Tr-En} } & & \multicolumn{3}{c}{\textbf{Tr-En-morph}} \\
 \cmidrule{3-5}  \cmidrule{7-9} \cmidrule{11-13}
&Embed size & 64 & 128 & 256 & & 64 & 128 & 256 & & 64 & 128 & 256 \\
&Model size (M) & $8\pm 1$ & $11\pm 1$ & $17\pm 1$ && $11\pm 1$ &$17\pm 1$ & $28\pm 1$ &&$13\pm 1$ &$21\pm 1$ & $36\pm 1$\\
\midrule
\multirow{ 3}{*}{4L}&BASE-4L & 28.97 & 29.99 & 30.43 && \textbf{19.80}  &20.26 & 20.99 && 18.90  & 18.81 & 20.08  \\
&DenseNMT-4L-1 & \textbf{30.11} & \textbf{30.80} &31.26 &&19.21 &20.08 & 21.36 &&18.83 & 20.16 & 21.43\\
&DenseNMT-4L-2 &29.77 & 30.01 &\textbf{31.40} && 19.59 & \textbf{20.86} & \textbf{21.48} && \textbf{19.04} & \textbf{20.19} & \textbf{21.57}\\
\midrule
\multirow{ 3}{*}{8L}&BASE-8L &30.15 & 30.91  &31.51  &&20.40  &21.60   &21.92  && 20.21 &20.76   & 22.62 \\ 
&DenseNMT-8L-1 &\textbf{30.91} & \textbf{31.54}  & 32.08 && 21.82  & \textbf{22.20}  &23.20 && 21.20 & 21.73  & 22.60  \\ 
&DenseNMT-8L-2 &30.70 & 31.17  & \textbf{32.26} && \textbf{21.93} & 21.98  &\textbf{23.25}  &&\textbf{21.73} & \textbf{22.44}  & \textbf{23.45} \\ 
\bottomrule
\end{tabular}
\caption{BLEU score on IWSLT German-English and Turkish-English translation tasks. We compare models using different embedding sizes, and keep the model size consistent within each column.}
\label{tab:embed}
\end{table*}

\begin{figure}[t]
\centering 
\captionsetup{font=small}
\includegraphics[width=0.4\textwidth]{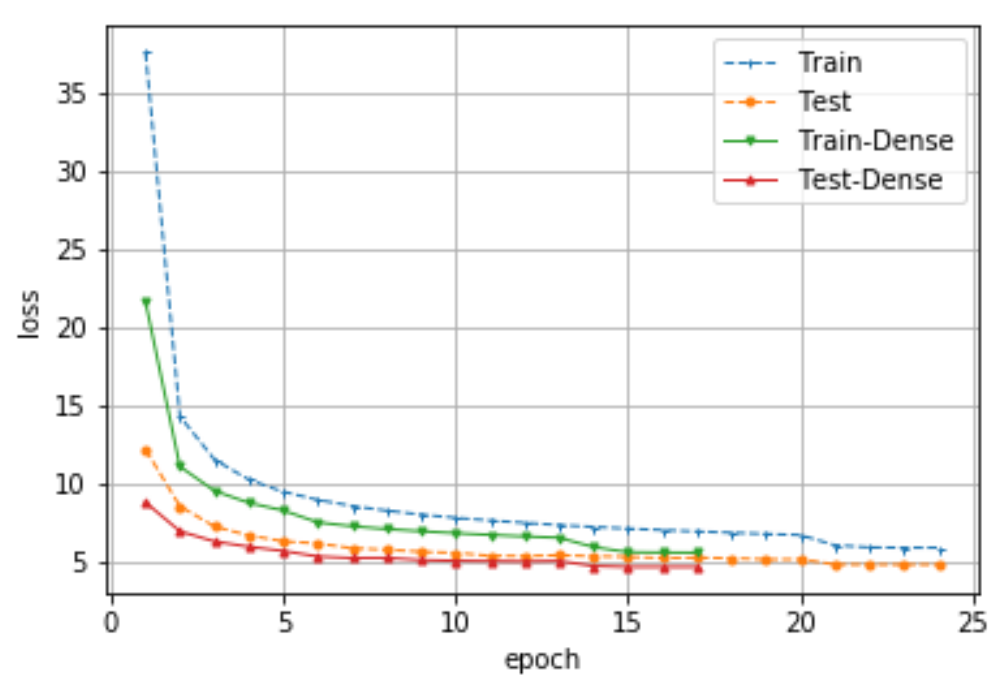}
\caption{Training curve and test curve comparison on WMT14 English-German translation task.}
\label{fig:curve_ende}
\end{figure}

\subsection{Training curve}
We first show that DenseNMT helps information flow more efficiently by presenting the training loss curve. 
All hyperparameters are fixed in each plot, only the models are different. 
In Figure \ref{fig:curve-appendix}, the loss curves for both training and dev sets (before entering the finetuning period) are provided for De-En, Tr-En and Tr-En-morph. 
For clarity, we compare \emph{DenseNMT-4L-2} with \emph{BASE-4L}. 
We observe that DenseNMT models are consistently better than residual-connected models, since their loss curves are always below those of the baseline models. 
The effect is more obvious  on the WMT14 English-German dataset. We rerun the best model provided by~\cite{convs2s} and compare with our model. 
In Figure \ref{fig:curve_ende}, where train/test loss curve are provided, DenseNMT-En-De-15 reaches the same level of loss and starts finetuning (validation loss starts to increase) at epoch 13, which is 35\% faster than the baseline. 

Adding dense connections changes the architecture, and would slightly influence training speed. For the WMT14 En-De experiments, the computing time for both DenseNMT and the baseline (with similar number of parameters and same batch size) tested on single M40 GPU card are 1571 and 1710 word/s, respectively. While adding dense connections influences the per-iteration training slightly (8.1\% reduction of speed), it uses many fewer epochs, and achieves a better BLEU score. In terms of training time, DenseNMT uses 29.3\%(before finetuning)/22.9\%(total) less time than the baseline.

\subsection{DenseNMT improves accuracy with similar architectures and model sizes}

\begin{table}[htbp]
\small
\centering
\captionsetup{font=small}
\begin{tabular}{lccc}
\toprule
 & \textbf{De-En}  & \textbf{Tr-En} & \textbf{Tr-En-morph}\\
\midrule
BASE & 30.43 & 20.99 & 20.08 \\
DenseENC-4L & 30.72 & 21.32 & 21.24   \\ 
DenseDEC-4L & 31.23 & 21.04 & 21.06  \\
DenseAtt-4L & 31.05 & 21.35 & 21.08  \\
DenseNMT-4L-1 & 31.26 & 21.36 & 21.43 \\ 
DenseNMT-4L-2 & 31.40 & 21.48 & 21.57\\ 
\bottomrule
\end{tabular}
\caption{ Ablation study for encoder block, decoder block, and attention block in DenseNMT.}
\label{tab:ablation}
\end{table}

Table \ref{tab:embed} shows the results for De-En, Tr-En, Tr-En-morph datasets, where the best accuracy for models with the same depth and of similar sizes are marked in boldface. In almost all genres, DenseNMT models are significantly better than the baselines. 
With embedding size 256, where all models achieve their best scores, DenseNMT outperforms baselines by 0.7-1.0 BLEU on De-En, 0.5-1.3 BLEU on Tr-En, 0.8-1.5 BLEU on Tr-En-morph. 
We observe significant gain using other embedding sizes as well.

Furthermore, in Table \ref{tab:ablation}, we investigate DenseNMT models through ablation study. 
In order to make the comparison fair, six models listed have roughly the same number of parameters. 
On De-En, Tr-En and Tr-En-morph, we see improvement by making the encoder dense, making the decoder dense, and making the attention dense. Fully dense-connected model \emph{DenseNMT-4L-1} further improves the translation accuracy. By allowing more flexibility in dense attention, \emph{DenseNMT-4L-2} provides the highest BLEU scores for all three experiments.  

From the experiments, 
we have seen that enlarging the information flow in the attention block benefits the models. The dense attention block provides multi-layer information transmission from the encoder to the decoder, and to the output as well. Meanwhile, as shown by the ablation study, the dense-connected encoder and decoder both give more powerful representations than the residual-connected counterparts. 
As a result, the integration of the three parts improve the accuracy significantly.

\begin{table*}[htbp]
\small
\centering
\captionsetup{font=small}
\begin{tabular}{lccccc}
\toprule
& \multicolumn{5}{c}{\textbf{Test Set}} \\
 & \textbf{tst2011} & \textbf{tst2012} & \textbf{tst2013} & \textbf{tst2014} & \textbf{total}\\
\midrule
RNN \cite{gulcehre2015using} & 18.40 & 18.77 & 19.86 & 18.64 & / \\
\midrule
BASE &21.66 &22.45 &23.76 &22.59 & 22.62 \\
DenseNMT-8L-2 &22.52 &23.81 &23.91 &23.68 & 23.45 \\
DenseNMT-8L-2(embed 256, hid 128) &23.33 & 24.65 & 24.92 & 24.54& 24.36 \\
\bottomrule
\end{tabular}
\caption{Accuracy on Turkish-English translation task in terms of BLEU score.}
\label{tab:tr-en}
\end{table*}

\subsection{DenseNMT with smaller embedding size}
From Table \ref{tab:embed}, we also observe that DenseNMT performs better with small embedding sizes compared to residual-connected models with regular embedding size. For example, on Tr-En model,  the $8$-layer \emph{DenseNMT-8L-2} model with embedding size $64$ matches the BLEU score of the $8$-layer BASE model with embedding size 256, while the number of parameter  of the former one is only $40\%$  of the later one. 
In all genres, DenseNMT model with embedding size $128$ is comparable or even better than the baseline model with embedding size $256$. 

While overlarge embedding sizes hurt accuracy because of overfitting issues, smaller sizes are not preferable because of insufficient representation power. However, our dense models show that with better model design, the embedding information can be well concentrated on fewer dimensions, e.g., 64. This is extremely helpful when building models on mobile and small devices where the model size is critical. While there are other works that stress the efficiency issue by using techniques such as separable convolution~\cite{slicenet}, and shared embedding~\cite{transformer}, our DenseNMT framework is orthogonal to those approaches. 
We believe that other techniques would produce more efficient models through combining with our DenseNMT framework. 

\begin{table}[htbp]
\small
\centering
\captionsetup{font=small}
\begin{tabular}{lcccc}
\toprule
 & \textbf{Greedy} & \textbf{Beam}\\
\midrule
MIXER~\cite{ranzato2015sequence} & 20.73 & 21.83  \\
AC~\cite{bahdanau2016actor} & 27.49& 28.53\\
NPMT~\cite{msrphrase} &27.83 & 28.96\\
NPMT+LM~\cite{msrphrase} & / & 29.16 \\
\midrule 
DenseNMT-8L-2 (word) & 29.11 & 30.33 \\
DenseNMT-8L-1 (BPE) &30.50  & 32.08 \\
DenseNMT-8L-2 (BPE)& 30.80 & 32.26 \\
\bottomrule
\end{tabular}
\caption{ Accuracy on IWSLT14 German-English translation task in terms of BLEU score.}
\label{tab:de-en}
\end{table}

\subsection{DenseNMT compares with state-of-the-art results}
For the IWSLT14 German-English dataset, we compare with the best results reported from literatures. To be consistent with prior works, we also provide results using our model directly on the dataset without BPE preprocessing. As shown in Table \ref{tab:de-en}, DenseNMT outperforms the phrase-structure based network NPMT~\cite{msrphrase} (with beam size 10) by 1.2 BLEU, using a smaller beam size, and outperforms the actor-critic method based algorithm~\cite{bahdanau2016actor} by 2.8 BLEU. For reference, our model trained on the BPE preprocessed dataset achieves 32.26 BLEU, which is 1.93 BLEU higher than our word-based model. 
For Turkish-English task, we compare with~\cite{gulcehre2015using} which uses the same morphology preprocessing as our Tr-En-morph. As shown in Table \ref{tab:tr-en}, our baseline is higher than the previous result, and we further achieve new benchmark result with 24.36 BLEU average score.
For WMT14 English-German, from Table \ref{tab:en-de}, we can see that DenseNMT outperforms ConvS2S model by 0.36 BLEU score using 35\% fewer training iterations and 20\% fewer parameters. We also compare with another convolution based NMT model: SliceNet~\cite{slicenet}, which explores depthwise separable convolution architectures. SliceNet-Full matches our result, and SliceNet-Super outperforms by 0.58 BLEU score. However, both models have 2.2x more parameters than our model. We expect DenseNMT structure could help improve their performance as well.

\begin{table}[htbp]
\centering
\small
\captionsetup{font=small}
\begin{tabular}{lcccc}
\toprule
 & \textbf{BLEU score}\\
\midrule
GNMT {\tiny~\cite{gnmt}} & 24.61 \\
ConvS2S {\tiny~\cite{convs2s}} & 25.16  \\
SliceNet-Full {\tiny~\cite{slicenet}} & 25.5  \\
SliceNet-Super {\tiny~\cite{slicenet}} & 26.1  \\
\midrule
DenseNMT-En-De-15 & 25.52   \\
\bottomrule
\end{tabular}
\caption{Accuracy on WMT14 English-German translation task in terms of BLEU score.}
\label{tab:en-de}
\end{table}

\section{Conclusion}
In this work, we have proposed DenseNMT as a dense-connection framework for translation tasks, which uses the information from embeddings more efficiently, and passes abundant information from the encoder side to the decoder side. Our experiments have shown that DenseNMT
is able to speed up the information flow and improve translation accuracy. For the future work, we will combine dense connections with other deep architectures, such as RNNs~\cite{gnmt} and self-attention networks~\cite{transformer}.

\bibliography{naaclhlt2018}
\bibliographystyle{acl_natbib}

\end{document}